# Foresight - Generative Pretrained Transformer (GPT) for Modelling of Patient Timelines using EHRs

*Zeljko Kraljevic[1,6], Dan Bean[1,6], Anthony Shek[2,3], Rebecca Bendayan[1,6], Harry Hemingway[4,5,7], Joshua Au Yeung[2,3], Alexander Deng[3], Alfred Baston[3], Jack Ross[3], Esther Idowu[3], James T Teo\*[2,3] and Richard JB Dobson\*[1,4,5,6,7]*

*Affiliations:*
*1 Department of Biostatistics and Health Informatics, Institute of Psychiatry, Psychology and Neuroscience, King's College London, London, U.K. London*
*2 Department of Neurology, King's College Hospital NHS Foundation Trust, Denmark Hill, London, London, U.K.*
*3 Guy's and St Thomas' NHS Foundation Trust*
*4 Health Data Research UK London, University College London, London, U.K.*
*5 Institute of Health Informatics, University College London, London, UK*
*6 NIHR Biomedical Research Centre at South London and Maudsley NHS Foundation Trust and King's College London, London, U.K.*
*7 NIHR Biomedical Research Centre at University College London Hospitals NHS Foundation Trust, London, UK*

**Evidence before this study:** We reviewed published evidence using Google Scholar and PubMed for studies using transformer-based models for forecasting patient timelines. We used the terms ("transformer" OR "bert" OR "generative pretrained transformer") AND ("forecasting" OR "temporal modelling" OR "trajectory") AND ("ehr" OR "health records" OR "medical records" OR "healthcare" OR "medicine" OR "patients" OR "hospital" OR "clinical"), the scope was anywhere in the text, published in 2018 or later. We found many CoVid-19 studies, or studies that focus on a specific biomedical concept or set of concepts. A few studies focus on forecasting a wider range of biomedical concepts, but still require structured data, work with specific timeframes or can only forecast one step into the future.

**Added value of this study:** Foresight can use unstructured and structured data, can work with different temporal resolutions (e.g. day, week, month) and because it is a generative model, in theory, it can simulate the patient's journey until death. Foresight was tested across hospitals, covering both physical and mental health, and 5 clinicians performed an independent test by simulating patients and outcomes. The tests were not focused on specific disorders or biomedical concepts but cover a broad range of concepts from the SNOMED ontology with 18 different concept types (e.g. Disorders, Substances, Findings and Procedures).

**Implications of all the available evidence:** Foresight is a powerful tool for forecasting medical concepts with application for medical education, simulation of patient journeys and causal inference research. Being derived from real-world data and modelling historical common practice, it is not expected to be perfectly consistent with contemporary recommended best practice clinical guidelines, so it should not be used for clinical decision support in its current form. As an iterative model, Foresight will improve with more real-world data and improved language processing.

\*The authors contributed equally

***Background:*** Electronic Health Records hold detailed longitudinal information about each patient's health status and general clinical history, a large portion of which is stored within the unstructured text. Existing approaches focus mostly on structured data and a subset of single-domain outcomes. We explore how temporal modelling of patients from free text and structured data, using deep generative transformers can be used to forecast a wide range of future disorders, substances, procedures or findings.

***Methods***: We present Foresight, a novel transformer-based pipeline that uses named entity recognition and linking tools to convert document text into structured, coded concepts, followed by providing probabilistic forecasts for future medical events such as disorders, substances, procedures and findings. We processed the entire free-text portion from three different hospital datasets totalling 811336 patients covering both physical and mental health.

***Findings***: On tests in two UK hospitals (King's College Hospital, South London and Maudsley) and the US MIMIC-III dataset precision@10 0·68, 0·76 and 0·88 was achieved for forecasting the next disorder in a patient timeline, while precision@10 of 0·80, 0·81 and 0·91 was achieved for forecasting the next biomedical concept. Foresight was also validated on 34 synthetic patient timelines by five clinicians and achieved relevancy of 97% for the top forecasted candidate disorder. As a generative model, it can forecast follow-on biomedical concepts for as many steps as required.

***Interpretation:*** Foresight is a general-purpose model for biomedical concept modelling that can be used for real-world risk forecasting, virtual trials and clinical research to study the progression of disorders, simulate interventions and counterfactuals, and educational purposes.

***Funding***: Part of a programme of work that received funding from the NHS AI Lab, National Institute of Health Research BRC and Health Data Research UK. Infrastructure support from KCH, SLaM Biomedical Research Centre and the London AI Centre for Value-Based Healthcare.

**Introduction**

Electronic Health Records (EHRs) contain detailed, longitudinal information about patients' health status and clinical history, much of which is stored in unstructured clinical notes. Previous research on forecasting using EHRs has primarily focused on structured data within EHRs and has often been limited to forecasting specific outcomes within a specific time frame. However, structured datasets are not always available, and even when they are, they can provide a limited view of a patient's journey, as approximately 80% of patient data is found in free text(1,2). Many previous studies are building upon BERT(3). One example is BEHRT(4) which uses a limited subset of disorders (301 in total) available in the structured portion of EHRs. BEHRT is limited to forecasts of disorders occurring in the next patient hospital visit or a specific predefined time frame, consequently requiring that the information is grouped into patient visits. In addition, we note that the approach is a multi-label approach, which can cause difficulties as the number of concepts to be forecasted increases. Another example is G-BERT(5) the inputs for this model are all single-visit samples, which are insufficient to capture long-term contextual information in the EHR. Like in BEHRT, only structured data is used. Lastly, Med-BERT(6) is trained on structured diagnosis data, coded using the International Classification of Diseases (ICD). The model is not directly trained on the target task of forecasting a new disorder but is fine-tuned after the standard Masked Language Modelling (MLM) task. The model is limited to ICD-10 codes and evaluated on a small subset of disorders which may be insufficient for estimating general performance. Apart from BERT-based models, we also note Long Short Term Memory (LSTM) models, like the one proposed by Steinberg et al. LM-LSTM(7). Like other models, they only use structured data and fine-tune their model to forecast limited future events.

We use the unstructured and structured data within the EHR to train a novel model, Foresight, for disorder and more generally biomedical concept forecasting. This work, to some extent, follows the approach outlined in GPTv3(8) where different tasks are implicit in the dataset; for example, one GPTv3 model can generate HTML code, answer questions, write stories and much more without any fine-tuning. We see the same in Foresight because the same model can be used to forecast the risk of disorders, offer differential diagnosis, suggest substances to be used and more. We test the model across multiple hospitals covering both physical and mental health and make it publicly available via a web application ( https://foresight.sites.er.kcl.ac.uk/ ).

## Methods

### Overview of the Foresight Pipeline

The Foresight pipeline (Figure 1) has four main components 1) CogStack(1), for data retrieval and the first step of data pre-processing; 2) MedCAT(9), for structuring of the free text information from EHRs; 3) Foresight Core, the deep learning model for biomedical concept modelling; and 4) Foresight web-app for interacting with the trained model.

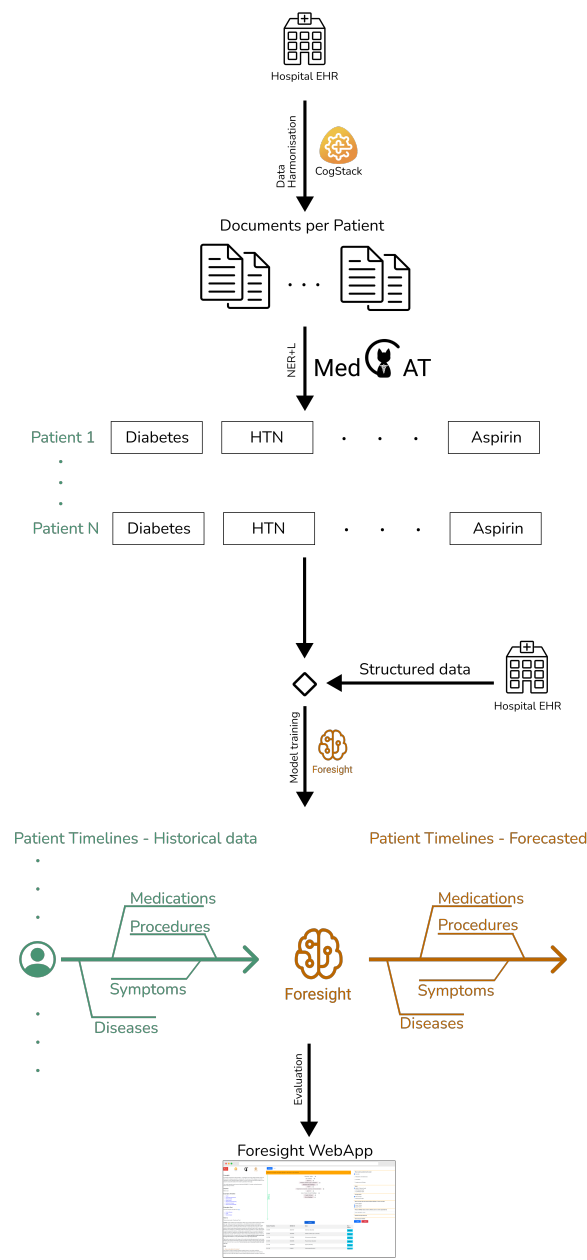

Figure 1. The foresight pipeline.

**Data Collection**

We used three datasets to train/test Foresight: 1) King's College Hospital (KCH) NHS Foundation Trust - all available free text from EHRs from 1999 to January 2021); 2) South London and Maudsley (SLaM) NHS Foundation Trust - all available free text for patients with a serious mental illness diagnosed prior to August 2019. SLaM is one of Europe's largest providers of secondary mental healthcare, serving a geographical catchment of approximately 1·32 million residents, and providing almost complete coverage of secondary mental healthcare provision to all age groups; 3) MIMIC-III - a publicly available dataset developed by the MIT Lab for Computational Physiology, consisting of data associated with patients who stayed in critical care units of the Beth Israel Deaconess Medical Center between 2001 and 2012.

*KCH Dataset*

At KCH we collected a total of 18436789 documents from 1459802 patients (both inpatients and outpatients) from the Allscripts Sunrise EHR using the CogStack platform(1). We retained document types known to be clinically information-rich and removed documents with OCR issues, incomplete triage checklists, questionnaires and forms. Documents have a timestamp representing the time they was written. Some documents were continuous, meaning more information was added to them over time (e.g. clinical notes). These were split into fragments, each containing a time of writing.

The project operated under London Southeast Research Ethics Committee (reference 18/LO/2048) approval granted to the King's Electronic Records Research Interface (KERRI); specific approval for the use of NLP on unstructured clinical text for extraction of standardised biomedical Codes for patient record structuring was reviewed with expert patient input on a patient-led committee with Caldicott Guardian oversight and granted Feb 2020.

*SLaM and MIMIC-III Datasets*

Both SLaM and MIMIC-III datasets were already organised and cleaned. At SLaM, we collected 14995092 documents from 27929 patients with a serious mental illness diagnosis using the CRIS system(10). While the number of documents at SLaM is comparable to KCH, the documents at SLaM are significantly shorter. For MIMIC-III, we used all available free text from clinical notes totalling 2083179 documents from 46520 patients.

This project was approved by the CRIS Oversight Committee, responsible for ensuring all research applications comply with ethical and legal guidelines.

**Named Entity Recognition and Linking**

The Medical Concept Annotation Toolkit (MedCAT) was used to extract biomedical concepts from free text and link them to the SNOMED-CT UK Clinical Edition and Drug Extension (hereafter referred to as SNOMED) concept database. MedCAT uses self-supervised learning to train a Named Entity Recognition and Linking (NER+L) model for any concept database (here SNOMED). MedCAT also supports concept contextualisation e.g. Negation detection (is the concept negated or not), which was important for this work as we were only interested in biomedical concepts from free text that are not negated and that are related to the patient. To train and validate MedCAT we manually annotated 17282 concept mentions from 698 randomly sampled documents from the full KCH dataset. The annotations were done by clinicians using MedCATtrainer(11) and were then used to fine-tune the base MedCAT model. The NER+L models were finetuned with a high precision bias, this was done due to the high level of redundancy in real-world health record data(12), so correct detection was more important as intrinsic redundancies make up for the occasionally missed concept.
We trained two new MedCAT contextualisation models (experiencer and negation) on the 17282 annotations. We then combined the contextualization and NER+L MedCAT models and annotated the entire datasets at KCH/MIMIC/SLaM. To test the patient-level Precision, 100 patients from each dataset were randomly sampled, and from each one we randomly picked a concept and manually verified whether it was correctly or incorrectly detected. We used the >1 occurrences rule, meaning a concept is only considered if it appears at least two times for a patient. MedCAT was not used with the full SNOMED ontology but a subset including Disorder, Substance, Finding and Procedure concepts (full list in Appendix 1). We ended up with 195416 different biomedical concepts from SNOMED.

Once the concepts were extracted, we removed all concepts that occurred <100 times in the whole dataset (to remove rare concepts that could identify patients) and grouped them by patient and organised into a timeline (Table 1 and 2). The datasets were split randomly into a train set (95%) and a test set (5%). We improved the quality and enriched the timeline by: 1) Keeping a concept that appeared at least twice in the patient's timeline, increasing the precision of our NER+L tool at the cost of recall; 2) Prepending age, sex and ethnicity to the timelines; 3) Adding a token denoting patient's age changes between concepts; 4) Removed concepts that are parents of concepts already in the timeline (i.e. in the past) to denoise the timeline, as in most cases, a parent of an existing concept does not bring any new information; 5) Appending <patient has died> token if the patient had died (only in our largest dataset at KCH); 6) Splitting the timeline into fragments of length N (also known as buckets, set to 1 day in our case) and removing duplicates within each fragment; 7) appending <SEP> tokens between fragments; and 8) Splitting timelines longer than L (L = 256 concepts in our case) and removing if shorter than 10 concepts; lastly

|  | KCH | | SLaM | | MIMIC-III | |
|---|---|---|---|---|---|---|
|  | Train | Test | Train | Test | Train | Test |
| Patients | 710194 | 37301 | 21910 | 1155 | 38749 | 2027 |
| Patients by Ethnicity | | | | | | |
| Asian | 34616 (5%) | 1764 (5%) | 1405 (6%) | 63 (6%) | 1031 (3%) | 58 (3%) |
| Black | 131216 (18%) | 6980 (19%) | 4822 (22%) | 281 (24%) | 3127 (8%) | 146 (7%) |
| Mixed | 8484 (1%) | 441 (1%) | 572 (3%) | 28 (2%) | 82 (0%) | 6 (0%) |
| Other | 34434 (5%) | 1798 (5%) | 4167 (19%) | 213 (19%) | 2428 (6%) | 120 (6%) |
| Unknown | 154132 (22%) | 8071 (21%) | 1150 (5%) | 48 (4%) | 4581 (12%) | 263 (13%) |
| White | 347312 (49%) | 18247 (49%) | 9794 (45%) | 522 (45%) | 27500 (71%) | 1434 (71%) |
| Patients by Sex | | | | | | |
| Female | 381155 (54%) | 19873 (53%) | 10054 (46%) | 544 (47%) | 16869 (44%) | 868 (43%) |
| Male | 328866 (46%) | 17422 (47%) | 11777 (54%) | 607 (53%) | 21880 (56%) | 1159 (57%) |
| Unknown | 173 (0%) | 6 (0%) | 79 (0%) | 4 (0%) | 0 (0%) | 0 (0%) |
| Patients by Age | | | | | | |
| 0-18 | 119297 (14%) | 6402 (14%) | 1437 (4%) | 81 (4%) | 3639 (9%) | 187 (9%) |
| 18-30 | 122137 (14%) | 6435 (15%) | 7372 (21%) | 378 (20%) | 1727 (4%) | 90 (4%) |
| 30-41 | 138706 (16%) | 7232 (17%) | 9009 (26%) | 500 (27%) | 2355 (6%) | 105 (5%) |
| 41-50 | 120187 (15%) | 6390 (14%) | 7283 (21%) | 393 (21%) | 3895 (10%) | 207 (10%) |
| 51-64 | 161799 (19%) | 8391 (19%) | 6044 (18%) | 345 (19%) | 9481 (24%) | 496 (24%) |
| 64+ | 183423 (22%) | 9489 (21%) | 3346 (10%) | 170 (9%) | 18648 (47%) | 990 (48%) |

Table 1. Selected characteristics from KCH, SLaM and MIMIC-III after preprocessing and timeline creation. For the *number of patients by age,* we multi-counted if one patient had data which spanned across more than one age group and the percentages in this case refer to the number of timelines instead of patients.

| | KCH | SLaM | MIMIC-III |
|---|---|---|---|
| Annotations (Unique) | 56736380 (10512) | 8958567 (2182) | 5046821 (2951) |
| Annotations per Semantic Type - Total (Unique) | | | |
| Disorder | 19003851 (5632) | 1743625 (674) | 2212841 (1376) |
| Substance | 12191307 (1185) | 2245368 (255) | 891764 (472) |
| Finding | 17282165 (2868) | 4747863 (929) | 1422286 (755) |
| Procedure | 3056147 (63) | 45189 (26) | 181254 (34) |

Table 2. Four common clinically relevant semantic types after dataset annotation from KCH, SLaM and MIMIC-III. Everything is calculated after data preprocessing and timeline formation.

**Foresight - biomedical concept forecasting**

Foresight is a transformer-based pipeline for modelling biomedical concepts from clinical narratives (Figure 2). It is built on top of the Generative Pretrained Transformer v2(14) architecture allowing causal language modelling (CLM). EHR data is sequentially ordered in time, and this sequential order is important(15). As such Masked Language Modelling (MLM) approaches like BERT(3) were not a good fit because when forecasting the masked token, BERT models can also look into the future (i.e. they are bi-directional). Formally the task at hand can be defined as given a corpus of patients $U = \{u_1, u_2, u_3, \ldots\}$ where each patient is defined as a sequence of tokens $u_i = \{w_1, w_2, w_3, \ldots\}$ and each token is medically relevant and temporally defined piece of patient data, our objective is the standard language modelling objective:

$$L(U) = \sum_i \sum_j \log P(w^i_j \mid w^i_{j-1}, w^i_{j-2}, \ldots w^i_0) \quad Eq.1$$

In this work each of the tokens $w_i$ represents a biomedical concept such as disorder, substance or finding (full list in Appendix 1) or patient demographics such as age, gender and ethnicity.

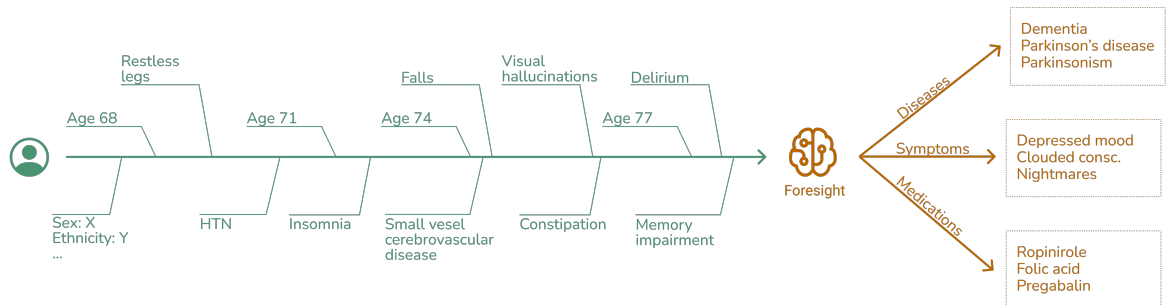

Figure 2. The left portion of the timeline represents the existing/historical data for a patient and the right portion are forecasts from Foresight for different biomedical concept types.

To find the optimal training hyperparameters for the Foresight transformer, we used Population Based Training(16) at KCH on the validation set (5% of the train set), the best result was achieved with n_layers=16, n_attention_heads=16, embedding_dim=512, weight_decay=1e-2, lr=3·14e-4, batch_size=32, and warmup_ratio=0·01, scheduler used was linear and we run the training for 10 epochs.

**Foresight web app**

To enable easier interaction with the model, the Foresight web app is available at https://foresight.sites.er.kcl.ac.uk/ . It can be used to evaluate the model for forecasting biomedical concepts by manually creating a patient timeline or loading an existing timeline. To understand why a certain concept was forecasted, we have added a gradient-based saliency method(17) to the web app allowing calculation and visualisation of concept importance for forecasting the next concept in sequence. The web app is also integrated with MedCAT, to enable analysis of unstructured text as input.

**Metrics**

The performance of models is measured using custom metrics that are an extension of the standard precision (TP / TP + FP) and recall (TP / TP + FN) aiming to replicate what the model will be used for but also consider the limitations of the training data.

At each point in a patient's timeline, the model forecasts the next concept. When measuring precision/recall, if the model forecasts that concept X will occur next while it should be concept Y, this forecast is not necessarily wrong. Several factors can influence what exactly is the next concept including a) The way the patient data is entered can significantly change the order of concepts in our timeline (albeit only on a short-time scale); b) Delayed diagnosis; c) Order of how the concept data is recorded in the EHR; and d) Concepts like chronic disorders, that do not have a precise starting point in a patient timeline but can appear a year before/after the real onset. Because of this, when determining whether the forecast is correct, we must evaluate forecasted concepts appearing in a certain time range. We define the following time ranges: 30 days, 1 year, and infinity (meaning all remaining data for a specific patient). For example, if we take the 30-day time range, a forecast is considered correct if the forecasted concept appears anywhere in that 30-day time range. We did not change the task at hand, and the model is still forecasting the next concept in the timeline, only the way we calculate metrics is modified.

As the model can be used for risk or diagnosis forecast, we are interested in how likely one of the top N forecasts is correct or, in other words, will appear in a patient's future. We used top-k @ {1, 5, 10}

To prevent the model from always forecasting the commonest group of concepts every forecasted concept must match the 'type' of the ground truth concept at that position in the timeline. For example, if for a patient, the next concept in a timeline is 'Diabetes Mellitus (disorder)' the output of the model will be filtered to only concepts of the type 'disorder'.

Finally, for each concept, we keep track of whether the forecasted concept is a new concept or a recurring one in that patient's timeline. A new concept means it has never appeared in the patient's timeline until now, and recurring means it has appeared at least once in past. We also filter the model output so that the forecasts are new/recurring concepts depending on what the ground truth is.

## Results

**Named Entity Recognition and Linking**

For the extraction of biomedical concepts (disorders, substances, procedures and findings) from clinical text using MedCAT we achieved a precision of 0·9549, recall of 0·8077 and F1 of 0·8752 while the models without precision bias achieved precision of 0·9314, recall of 0·8959 and F1 of 0·9133. For the contextualization, the F1 scores were 0·9280 for Patient and 0·9490 for Negation. The patient level MedCAT precision for each datasets and concept tyle is in Table 3.

|  | Precision (True positive / False positive) | | |
| --- | --- | --- | --- |
|  | KCH | SLaM | MIMIC-III |
| Overall | 97% (97/3) | 98% (98/2) | 95% (95/5) |
| Disorder | 96% (48/2) | 100% (27/0) | 91% (44/3) |
| Substance | 95% (19/1) | 100% (26/0) | 94% (17/1) |
| Finding | 100% (24/0) | 96% (44/2) | 97% (31/1) |
| Procedure | NA (0/0) | 100% (1/0) | 100% (3/0) |
| Table 3. Patient-level precision for randomly selected 100 concepts from each of the three datasets. Each concept was required to have ≥2 occurrences in a timeline to be considered as present. | | | |

**Foresight - biomedical concept forecasting**

The average precision and recall for forecasting disorders in the largest dataset (KCH) is 0·55/0·47. Increasing the Time Range, in other words allowing for the forecasted concept to appear anywhere in a patient's future, increases the precision and recall to 0·64/0·54. Using Top-K of 10 instead of 1 (number of candidates we consider) results in a precision and recall of 0·84/0·76. Forecasting of recurring concepts works significantly better than forecasting of new concepts. Detailed results are in Table 4.

Regarding the size of the network, we find that adding more layers or increasing the heads x layers up to 32x32 did not make a difference, beyond which there was significant performance deterioration. Increasing the bucket size did not improve the performance; the model trained on bucket size of 1 day outperformed all other models trained on bucket sizes of 3, 7, 14, 30 and 365 days.

| | | | Precision/Recall | | | | | |
|---|---|---|---|---|---|---|---|---|
| Concept Type | Time Range (days) | Top-K | KCH New | KCH Recurring | SLaM New | SLaM Recurring | MIMIC-III New | MIMIC-III Recurring |
| All | 30 | 1 | 0.43/0.32 | 0.83/0.77 | 0.38/0.23 | 0.77/0.67 | 0.52/0.32 | 0.83/0.67 |
| All | 30 | 5 | 0.71/0.57 | 0.99/0.97 | 0.71/0.48 | 0.97/0.92 | 0.84/0.59 | 0.98/0.92 |
| All | 30 | 10 | **0.80/0.67** | **1.00/0.99** | **0.81/0.60** | **0.99/0.97** | **0.91/0.70** | **1.00/0.97** |
| All | 365 | 1 | 0.47/0.33 | 0.88/0.83 | 0.51/0.25 | 0.86/0.77 | 0.54/0.33 | 0.85/0.70 |
| All | inf | 1 | 0.51/0.34 | 0.89/0.86 | 0.56/0.26 | 0.88/0.80 | 0.55/0.33 | 0.86/0.70 |
| Disorders | 30 | 1 | 0.30/0.21 | 0.80/0.72 | 0.34/0.24 | 0.78/0.72 | 0.46/0.26 | 0.79/0.60 |
| Disorders | 30 | 5 | 0.57/0.43 | 0.98/0.96 | 0.65/0.49 | 0.98/0.96 | 0.79/0.51 | 0.98/0.89 |
| Disorders | 30 | 10 | **0.68/0.53** | **1.00/0.99** | **0.76/0.60** | **1.00/1.00** | **0.88/0.62** | **0.99/0.96** |
| Disorders | 365 | 1 | 0.35/0.23 | 0.87/0.81 | 0.44/0.26 | 0.86/0.80 | 0.49/0.26 | 0.83/0.64 |
| Disorders | inf | 1 | 0.38/0.23 | 0.89/0.84 | 0.49/0.27 | 0.87/0.83 | 0.50/0.26 | 0.84/0.65 |
| Findings | 30 | 1 | 0.41/0.26 | 0.77/0.70 | 0.39/0.19 | 0.72/0.59 | 0.52/0.29 | 0.83/0.66 |
| Findings | 30 | 5 | 0.70/0.51 | 0.98/0.95 | 0.72/0.42 | 0.95/0.87 | 0.85/0.58 | 0.99/0.93 |
| Findings | 30 | 10 | **0.80/0.63** | **1.00/0.99** | **0.82/0.55** | **0.99/0.95** | **0.92/0.70** | **1.00/0.98** |
| Findings | 365 | 1 | 0.46/0.27 | 0.82/0.76 | 0.55/0.22 | 0.82/0.71 | 0.54/0.29 | 0.85/0.67 |
| Findings | inf | 1 | 0.51/0.28 | 0.84/0.80 | 0.61/0.22 | 0.85/0.74 | 0.55/0.29 | 0.85/0.68 |
| Substances | 30 | 1 | 0.46/0.34 | 0.87/0.79 | 0.36/0.25 | 0.85/0.78 | 0.52/0.32 | 0.84/0.70 |
| Substances | 30 | 5 | 0.77/0.63 | 0.99/0.98 | 0.70/0.55 | 0.99/0.98 | 0.85/0.61 | 0.99/0.94 |
| Substances | 30 | 10 | **0.86/0.74** | **1.00/1.00** | **0.82/0.69** | **1.00/1.00** | **0.92/0.73** | **1.00/0.99** |
| Substances | 365 | 1 | 0.49/0.35 | 0.90/0.86 | 0.43/0.277 | 0.91/0.87 | 0.53/0.32 | 0.85/0.71 |
| Substances | inf | 1 | 0.52/0.36 | 0.91/0.89 | 0.46/0.28 | 0.92/0.89 | 0.54/0.32 | 0.85/0.71 |

| Procedures | 30 | 1 | 0.68/0.61 | 0.92/0.91 | 0.53/0.51 | 0.97/0.97 | 0.80/0.67 | 0.93/0.92 |
| --- | --- | --- | --- | --- | --- | --- | --- | --- |
| Procedures | 30 | 5 | 0.93/0.91 | 1.00/1.00 | 0.87/0.86 | 1.00/1.00 | 0.97/0.94 | 1.00/1.00 |
| Procedures | 30 | 10 | **0.97/0.97** | **1.00/1.00** | **0.96/0.96** | **1.00/1.00** | **0.99/0.99** | **1.00/1.00** |
| Procedures | 365 | 1 | 0.71/0.61 | 0.94/0.95 | 0.54/0.51 | 0.98/0.98 | 0.81/0.67 | 0.95/0.94 |
| Procedures | inf | 1 | 0.73/0.62 | 0.95/0.96 | 0.55/0.51 | 0.98/0.98 | 0.82/0.67 | 0.95/0.94 |

Table 4. Precision and Recall for next biomedical concept forecast.

*Clinical evaluation of the generated next biomedical concept*

5 clinicians produced 34 synthetic timelines for simulated scenarios similar to a 'clinical'; Each timeline was processed by Foresight (KCH model) and 5 forecasted Disorder concepts were presented back to the clinicians. In each example, the clinicians were asked to score the relevancy of each of the forecasted concepts. 'Relevancy of forecasted concepts' was chosen over 'Accuracy' as there were frequent disagreements on ground truth on which forecasted concept is most 'correct' (Table 5). Multiple answer relevancy is also more compatible with real-world clinical practice, which is geared towards concurrently considering and managing for multiple possible diagnoses, multiple investigations and multiple interventions rather than the classical "Single Best Answer" commonly used in UK medical examinations(18,19).

|  | 1 | 2 | 3 | 4 | 5 |
| --- | --- | --- | --- | --- | --- |
| Percentage of relevant concepts | 97% | 96% | 90% | 89% | 88% |

Table 5. Results of the manual clinician verification, the columns represent how many concept suggestions from the Foresight KCH model were evaluated, and how many of the suggestions were relevant.

Most concepts forecasted were relevant in sequence. The overall inter-annotator agreement is 86% (among the 5 clinicians). For cases where all clinicians agree, the percentage of relevant concepts out of the top 5 is 93%, where 4 out of 5 clinicians agree it is 81%, and where 3 out of 5 clinicians agree 62%. An example of a 'clinical vignette' with an error is presented in Figure 3 where 4 out of the 5 forecasted concepts ("normal pressure hydrocephalus", "hydrocephalus", "dementia" and "Alzheimer's disease") were relevant.

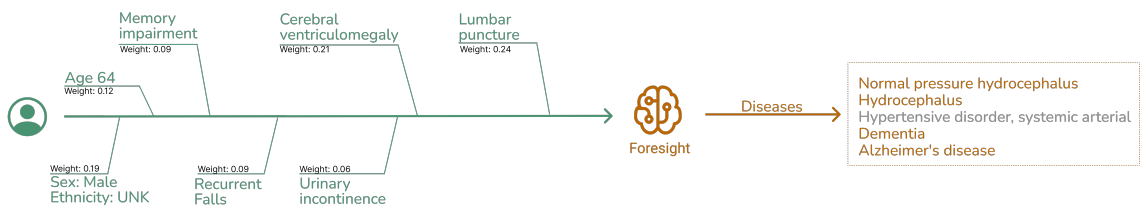

Figure 3. An example of a patient timeline with forecasted disorders. Saliency (weight) is shown for the first candidate - Normal Pressure Hydrocephalus. Irrelevant forecast in grey.

This is compatible with clinician heuristic reasoning to expect that the diagnosis was reached as a result of the last concept in the timeline - the "lumbar puncture" procedure (whether by CSF removal or molecular biomarkers) combined in the context of preceding symptoms. The single irrelevant concept of "hypertensive disorder, systemic arterial" failed to take this contextual cue and forecasted a diagnosis that though statistically very common in the age group was highly irrelevant in the context of the other concepts. As per Table 5, most concepts forecasted were relevant showing the contribution of the contextual attentional transformer mechanism in Foresight. For all other timelines and outputs, please review the Foresight repository at https://github.com/CogStack/Foresight .

*Examples of generating multiple synthetic concepts into the future*

Next we demonstrate that Foresight can forecast multiple concepts into the future and create whole patient timelines given a short prompt, in this case, 43-year-old, black, female (Figure 4). We use top-k sampling with k = 100 and generate timelines of 21 (6 base + 15 new) concepts.

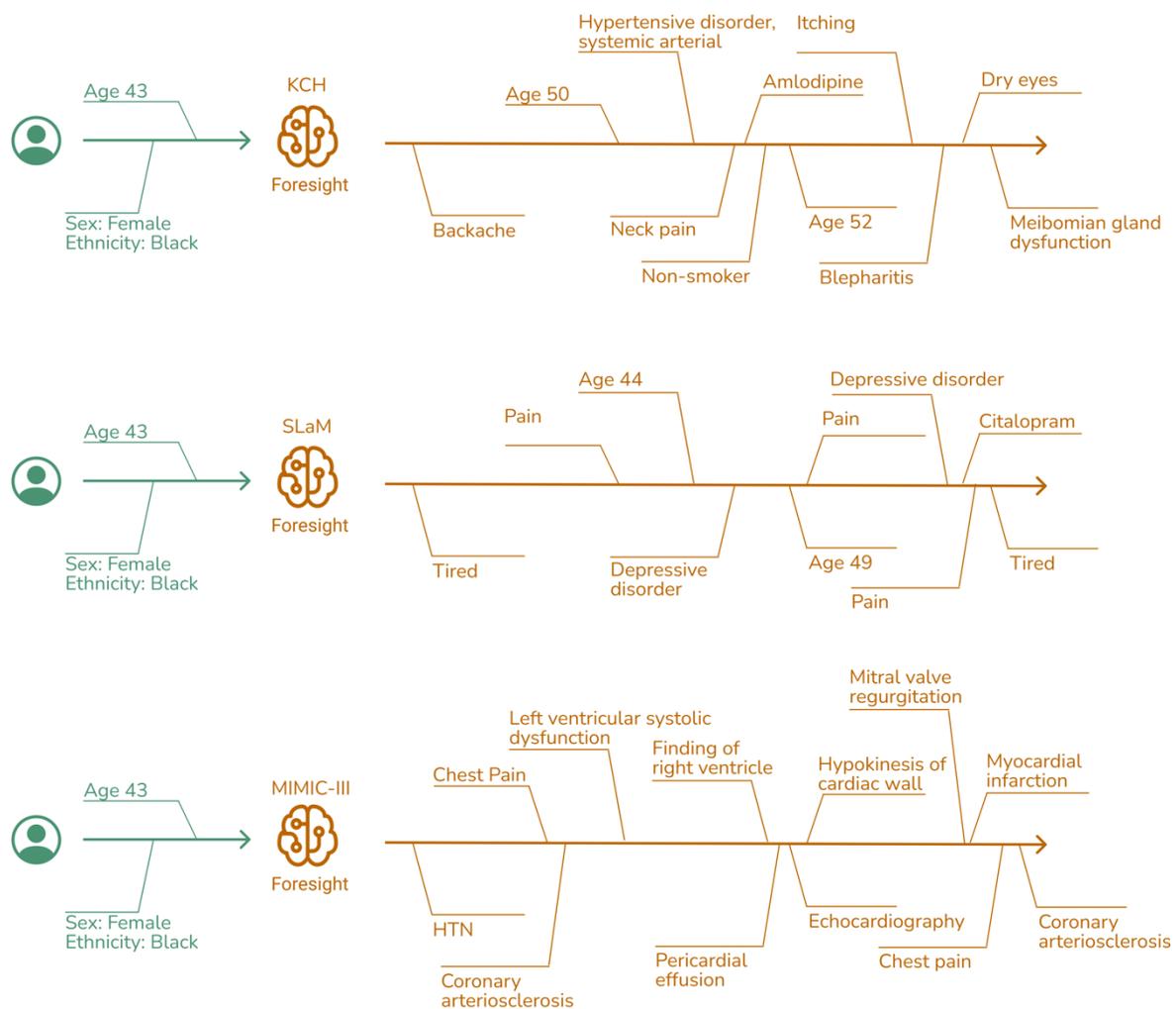

Figure 4. Generated synthetic timeline examples for input: 43-year-old, black, female (top – KCH model, middle – SLAM model, bottom - MIMIC-III model). The right side of the timelines (orange part) was forecasted by Foresight to simulate the medical future of a 43-year-old black female according to the 3 different models. The distances in the figures do not represent real temporal distances, only the order of concepts in the timelines is important.

## Discussion

We propose a novel deep-learning generative model of patient timelines within secondary care across mental and physical health, incorporating interoperable disorder, procedure, substance and finding concepts. Foresight is a system wide approach that targets entire hospitals and encompasses all patients together with any biomedical concepts (e.g. disorders) that can be found in both structured and unstructured parts of an EHR. One advantage of Foresight is that it can easily scale to more patients, hospital or disorders with minimal or no modifications, and the more data it receives the better it gets. As a generative model, Foresight is not limited to forecasting the next step, or patient episode; it can continue generating a patient timeline for any desired duration.

Foresight allows simulation of a patient future from single time-steps during a time-constrained inpatient episode all the way to a multi-year timeline of chronic conditions. This opens the door for research into "What if?" scenarios in Health Digital Twins. Digital Twins provide a way to estimate the impact of existing interventions on historical real-world data, beyond a purely dichotomous outcome incorporating how co-morbidities (both physical and mental health) may interact with each other and the primary outcome(20,21). Simulations with Foresight provide a route for counterfactual modelling to allow causal inference(22). Such a digital twin could also be used for medical education, where symptoms and medical histories are provided, and differential diagnoses and relevant investigations are quizzed against. This could also be played out into forecasted learning scenarios - the traditional 'clinical vignette' teaching method enhanced by deep learning for the digital era(23). Future work in this area should explore extended timeline simulation in more detail as well as improve on the generated timelines with, for example, a learning-to-rank model similar to how the CLIP(24) model works with DALL-E(25).

The ability to forecast diagnoses/ substances/ procedures is useful for education and exploring the impact in previous real-world practice. While there is a temptation to imagine the forecasted output to be used for clinical care or decision support - this is premature as Foresight is derived from historical common practice so would not be expected to be consistent with contemporary recommended best practice clinical guidelines. Clinical practice and disease patterns drift over time leading to treatments or diagnoses patterns that are era-specific - simulation of a patient with an upper respiratory tract infection in an Influenza-dominant era would be misguided in a Covid-dominant era. Availability of new treatments or interventions would also be under-represented in Foresight, and disease profile would be weighted to conditions and scenarios in secondary and tertiary care, i.e. it would be weighted towards more comorbidity as patients with lower complexity or early-stage conditions who are completely dealt with in primary care would be under-represented in our dataset(26).

Foresight prioritises *probability* of a concept over *urgency and impact* of a concept, while real-world clinical practice and heuristic clinical reasoning is often geared towards *high impact, high urgency, low probability* events over *low impact, low urgency, high probability* events. This can produce a scenario where forecasted concepts are common but irrelevant to the context, e.g. an elderly patient with a timeline culminating in "central crushing chest pain", is incidentally forecasted to have "cataracts" next, which is irrelevant to the more pressing scenario of the chest pain. This relevancy could be introduced through 'prompt engineering' to filter to only certain disease types or organ systems, types of medications, or to provide a separate relevancy signal. Finally, hallucinations are also well-described in Transformer-based generative models(27) including the recent ChatGPT, so such relevancy and mitigation systems would need to be built before any suitability for clinical decision support.

Due to the modular architecture of the system, the individual subcomponents can be improved or extended: (1) further tuning of the concept capture of the natural language processing; (2) inclusion of quantitative data like blood pressure measurements or blood test results; (3) expansion of dataset for greater coverage of Rare Disease while preserving privacy; and (4) representation of 'external knowledge' from published clinical guidelines, academic publications

We present a novel deep learning generative model of patients using EHRs that is composed of both natural language processing and a longitudinal forecasting, with broad utility across many healthcare domains. We anticipate further iterative improvements as all subcomponents are improvable. Foresight opens the door for digital health twins, synthetic dataset generation, real world risk forecasting, longitudinal research, emulation of virtual trials, medical education and more.

## Code and Data availability statement

- SLAM: Due to the confidential nature of free-text data, we are unable to make patient-level data available. CRIS was developed with extensive involvement from service users and adheres to strict governance frameworks managed by service users. It has passed a robust ethics approval process acutely attentive to the use of patient data. Specifically, this system was approved as a dataset for secondary data analysis on this basis by Oxfordshire Research Ethics Committee C (08/H06060/71). The data are deidentified and used in a data-secure format and all patients have the choice to opt-out of their anonymised data being used. Approval for data access can only be provided from the CRIS Oversight Committee at SLaM.
- KCH: Source patient-level dataset is not available for privacy reasons. The source dataset is described in the Health Data Research UK Innovation Gateway https://web.www.healthdatagateway.org/dataset/4e8d4fed-69d6-402c-bd0a-163c23d6b0ee with a wider timeframe (2010-2022).
- MIMIC-III data availability statement: MIMIC-III is available publicly at https://physionet.org
- Foresight: the code is available on GitHub at https://github.com/CogStack/Foresight and the web app can be accessed on https://foresight.sites.er.kcl.ac.uk/

## Disclaimer

This material includes SNOMED Clinical Terms® (SNOMED CT®) which is used by permission of the International Health Terminology Standards Development Organisation (IHTSDO). All rights reserved. SNOMED CT®, was originally created by The College of American Pathologists. "SNOMED" and "SNOMED CT" are registered trademarks of the IHTSDO.

## Author Contribution

Conceptualization: ZK, RJBD, DB, JTT, RB
Data curation: ZK, AS, JTT, JAY
Methodology: ZK
Supervision: RJBD, DB, RB
Clinical Validation: JTT, JAY, AD, AB, JR, EI
Software: ZK
Writing – original draft: ZK
Writing – review & editing: HH, JTT, RJBD, JTT, JAY, AD, AB, JR, EI, DB, RB

## Conflict of Interest

No conflict of interest.

## Acknowledgements

RD's work is supported by RJBD is supported by the following: (1) NIHR Biomedical Research Centre at South London and Maudsley NHS Foundation Trust and King's College London, London, UK; (2) Health Data

Research UK, which is funded by the UK Medical Research Council, Engineering and Physical Sciences Research Council, Economic and Social Research Council, Department of Health and Social Care (England), Chief Scientist Office of the Scottish Government Health and Social Care Directorates, Health and Social Care Research and Development Division (Welsh Government), Public Health Agency (Northern Ireland), British Heart Foundation and Wellcome Trust; (3) The BigData@Heart Consortium, funded by the Innovative Medicines Initiative-2 Joint Undertaking under grant agreement No. 116074. This Joint Undertaking receives support from the European Union's Horizon 2020 research and innovation programme and EFPIA; it is chaired by DE Grobbee and SD Anker, partnering with 20 academic and industry partners and ESC; (4) the National Institute for Health Research University College London Hospitals Biomedical Research Centre; (5) the National Institute for Health Research (NIHR) Biomedical Research Centre at South London and Maudsley NHS Foundation Trust and King's College London; (6) the UK Research and Innovation London Medical Imaging & Artificial Intelligence Centre for Value Based Healthcare; (7) the National Institute for Health Research (NIHR) Applied Research Collaboration South London (NIHR ARC South London) at King's College Hospital NHS Foundation Trust.

DB is funded by a UKRI Innovation Fellowship as part of Health Data Research UK MR/S00310X/1 (https://www.hdruk.ac.uk).

RB is funded in part by grant MR/R016372/1 for the King's College London MRC Skills Development Fellowship programme funded by the UK Medical Research Council (MRC, https://mrc.ukri.org) and by grant ISBRC-1215-20018 for the National Institute for Health Research (NIHR, https://www.nihr.ac.uk) Biomedical Research Centre at South London and Maudsley NHS Foundation Trust and King's College London.

THIS Institute.

This paper represents independent research part funded by the National Institute for Health Research (NIHR) Biomedical Research Centre at South London and Maudsley NHS Foundation Trust, The UK Research and Innovation London Medical Imaging & Artificial Intelligence Centre for Value Based Healthcare (AI4VBH); the National Institute for Health Research (NIHR) Applied Research Collaboration South London (NIHR ARC South London) and King's College London. The views expressed are those of the author(s) and not necessarily those of the NHS, MRC, NIHR or the Department of Health and Social Care. We thank the patient experts of the KERRI committee, Professor Irene Higginson, Professor Alastair Baker, Professor Jules Wendon, Professor Ajay Shah, Dan Persson and Damian Lewsley for their support.

# Appendix 1

A list of all concept types that were selected from the SNOMED ontology: Occupation; Disorder; Clinical drug; Tumour staging; Record artifact; Medicinal product form; Organism; Situation; Observable entity; Substance; Finding; Assessment scale; Medicinal product; Body structure; Physical object; Morphologic abnormality; Regime/Therapy; Product; Procedure.

# Appendix 2

| KCH | | | SLaM | | | MIMIC-III | | |
|---|---|---|---|---|---|---|---|---|
| Name | TP | FP |  | TP | FP |  | TP | FP |
| Fast Alcohol Screening Test (assessment scale)* | 51 | 0 | Cardiac pacemaker, device (physical object) | 22 | 0 | Care plan (record artifact) | 341 | 0 |
| Cellulitis of eyelid (disorder) | 45 | 0 | Conservative therapy (regime/therapy) | 12 | 0 | Cardiac pacemaker, device (physical object) | 166 | 0 |
| Deficiency of transaldolase (disorder) | 41 | 0 | Left kidney structure (body structure) | 9 | 0 | Anoxic encephalopathy (disorder) | 12 | 0 |
| Congenital disease (disorder) | 40 | 0 | Product containing antigen of whole cell pertussis and diphtheria toxoid and tetanus toxoid adsorbed (medicinal product) | 8 | 0 | Conservative therapy (regime/therapy) | 10 | 0 |
| Alpha-methylacyl-CoA racemase deficiency disorder (disorder) | 38 | 0 | Moderate pain (finding) | 6 | 0 | Product containing benzocaine in cutaneous dose form (medicinal product form) | 9 | 0 |
| Ichthyosis (disorder) | 38 | 0 | Sickle cell-hemoglobin SS disease (disorder) | 6 | 0 | Human immunodeficiency virus (organism) | 8 | 0 |
| McCune Albright syndrome (disorder) | 38 | 0 | Human immunodeficiency virus (organism) | 5 | 0 | Pseudocyst of pancreas (disorder) | 6 | 0 |
| Human immunodeficiency virus (organism) | 33 | 0 | Allergies and adverse reaction (record artifact) | 4 | 0 | Poor muscle tone (finding) | 5 | 0 |
| Polymyxin (substance) | 30 | 0 | Vasovagal syncope (disorder) | 2 | 0 | Status epilepticus (disorder) | 5 | 0 |
| Hepatitis C antibody test negative (finding) | 28 | 0 | Diurnal variation of mood (finding) | 2 | 0 | Fracture of pubic rami (disorder) | 5 | 0 |
| ⋮ | | | | | | | | |
| Sprain of ligament (disorder) | 145 | 1236 | Victim of neglect (finding) | 18 | 116 | Urinary tract infectious disease (disorder) | 33 | 107 |
| Radiating pain (finding) | 36 | 339 | Smartly dressed (finding) | 36 | 268 | Hyperlipidemia (disorder) | 75 | 250 |
| Varicella (disorder) | 39 | 410 | Omeprazole (substance) | 16 | 120 | Traumatic tear of skin (disorder) | 39 | 134 |
| Fibromyalgia (disorder) | 30 | 295 | Backache (finding) | 21 | 171 | Hypercholesterolemia (disorder) | 36 | 118 |
| Generally unwell (finding) | 18 | 192 | Non-smoker (finding) | 19 | 161 | Dry cough (finding) | 30 | 103 |
| Acne vulgaris (disorder) | 67 | 752 | Visual hallucinations (finding) | 16 | 124 | Depressive disorder (disorder) | 68 | 239 |
| Sprain of ankle (disorder) | 50 | 626 | Low blood pressure (disorder) | 14 | 130 | Left atrial abnormality (disorder) | 33 | 121 |
| Right bundle branch block (disorder) | 8 | 103 | Feeling mixed emotions (finding) | 21 | 255 | Oxycodone (substance) | 43 | 174 |
| Fracture of hand (disorder) | 15 | 228 | Lethargy (finding) | 9 | 108 | Abscess (disorder) | 27 | 109 |
| Open wound of hand (disorder) | 9 | 167 | Adequately dressed (finding) | 6 | 106 | Calculus in biliary tract (disorder) | 12 | 113 |

| | KCH | SLaM | MIMIC-III |
|---|---|---|---|

Table A1. Top and bottom 10 best/worst performing concepts with respect to precision, and the associated count in the test set. Precision from NEW concepts. TP - number of true positives and FP - number of false positives on the test set.
*These concepts are inaccuracies of disambiguation in the NER+L to be removed by further fine-tuning.

**Appendix 3**

| | KCH | | SLaM | | MIMIC-III | |
|---|---|---|---|---|---|---|
| | Train | Test | Train | Test | Train | Test |
| Mean Timeline Length in concepts (in years from first to last admission) | 75 (3.3) | 75 (3.3) | 387 (6.9) | 414 (7.3) | 123 (0.5) | 121 (0.5) |
| Mean Timeline Length by Ethnicity in concepts (in years from first to last admission) | | | | | | |
|   Asian | 80 (3.6) | 78 (3.5) | 361 (6.9) | 344 (7.4)* | 116 (0.5) | 102 (0.3)* |
|   Black | 77 (4.7) | 79 (4.6) | 524 (8.9) | 596 (9.2) | 141 (0.8) | 157 (0.7) |
|   Mixed | 55 (3.7) | 58 (3.6) | 516 (7.7) | 307 (6.9)* | 120 (0.5)* | 71 (0.1)* |
|   Other | 66 (3.2) | 65 (3.2) | 372 (6.3) | 367 (6.6) | 122 (0.5) | 131 (0.5) |
|   Unknown | 55 (2.1) | 55 (2.0) | 92 (1.6) | 58 (1.0)* | 91 (0.1) | 96 (0.1) |
|   White | 86 (3.4) | 85 (3.3) | 357 (6.7) | 382 (7.4) | 128 (0.5) | 122 (0.5) |
| Mean Timeline Length by Sex in concepts (in years from first to last admission) | | | | | | |
|   Female | 74 (3.5) | 74 (3.4) | 369 (6.8) | 394 (7.3) | 125 (0.5) | 123 (0.5) |
|   Male | 78 (3.2) | 77 (3.2) | 404 (7.0) | 434 (7.4) | 123 (0.5) | 119 (0.5) |
|   Unknown | 88 (1.5) | 16 (0.4)* | 238 (5.0)* | 109 (3.8)* | NA | NA |
| Mean Timeline Length by Age in concepts (in years | | | | | | |
| from first to last admission) | | | | | | |
| 0-18 | 47 (3.2) | 48 (3.2) | 237 (1.6) | 226 (1.6)* | 73 (0.1) | 31 (0.0) |
| 18-30 | 43 (2.8) | 42 (2.7) | 359 (3.6) | 373 (3.6) | 87 (0.3) | 73 (0.2)* |
| 30-41 | 50 (3.2) | 49 (3.2) | 405 (6.2) | 438 (6.7) | 103 (0.5) | 105 (0.5) |
| 41-50 | 67 (3.7) | 66 (3.5) | 414 (8.1) | 448 (8.0) | 119 (0.6) | 112 (0.6) |
| 51-64 | 87 (3.8) | 88 (3.8) | 432 (9.5) | 444 (10.2) | 126 (0.6) | 123 (0.6) |
| 64+ | 122 (3.4) | 121 (3.4) | 321 (7.7) | 365 (8.4) | 132 (0.6) | 128 (0.5) |
| Mean Number of Concepts of Certain Type per Timeline | | | | | | |
| Disorder | 25 | 25 | 75 | 81 | 54 | 53 |
| Substance | 16 | 16 | 97 | 102 | 21 | 21 |
| Finding | 23 | 23 | 205 | 221 | 35 | 34 |
| Procedure | 4 | 4 | 2 | 2 | 2 | 4 |

Table A2. Selected timeline characteristics from KCH, SLaM and MIMIC-III. For *mean timeline length by age,* we took the most recent age of a patient and used that to determine the age group. If a number is marked with an * it means the calculation was done on less than 100 timelines (patients).